%% file: main_ICCV.tex
\documentclass[10pt,twocolumn,letterpaper]{article}

\usepackage{iccv}              

\usepackage{microtype}
\usepackage{graphicx}
\usepackage{booktabs} 
\usepackage{multirow}  
\usepackage{amsmath} 
\usepackage{amssymb}
\usepackage{mathtools}
\usepackage{amsthm}
\usepackage{pifont}
\usepackage{color}
\usepackage{bbm}
\usepackage{array} 
\usepackage{makecell}
\usepackage{bm}
\usepackage{algorithm}
\usepackage{algorithmic}

\newcommand{\Cmat}[0]{\ensuremath{{\bf C}} }

\newcommand{\Tmat}[0]{\ensuremath{{\bf T}} }

\newcommand{\wv}[0]{\ensuremath{\boldsymbol{w}} }

\newcommand{\tableCellHeight}{1}
\newcommand{\tabstyle}[1]{
  \setlength{\tabcolsep}{#1}
  \renewcommand{\arraystretch}{\tableCellHeight}
  \centering
  \small
}

\input{preamble}

\definecolor{iccvblue}{rgb}{0.21,0.49,0.74}
\usepackage[pagebackref,breaklinks,colorlinks,allcolors=iccvblue]{hyperref}

\def\paperID{547} 
\def\confName{ICCV}
\def\confYear{2025}

\title{Dynamic Multimodal Prototype Learning in Vision-Language Models}

\author{
Xingyu Zhu\textsuperscript{1,2},
Shuo Wang\textsuperscript{1*},
Beier Zhu\textsuperscript{2},
Miaoge Li\textsuperscript{3},
Yunfan Li\textsuperscript{4},\\
Junfeng Fang\textsuperscript{5},
Zhicai Wang\textsuperscript{1},
Dongsheng Wang\textsuperscript{6},
Hanwang Zhang\textsuperscript{2} \\[0.5em]
\textsuperscript{1}University of Science and Technology of China, \
\textsuperscript{2}Nanyang Technological University, \\
\textsuperscript{3}The Hong Kong Polytechnic University, \
\textsuperscript{4}Sichuan University, \\
\textsuperscript{5}National University of Singapore, \
\textsuperscript{6}Shenzhen University \\
{\tt\small xyzhuxyz@mail.ustc.edu.cn}
}

\begin{document}
\maketitle
\let\thefootnote\relax\footnote{*Corresponding author.}
\input{section/0_abstract}
\input{section/1_introduction}

\input{section/2_related_work}
\input{section/3_method}

\input{section/4_experiments}
\input{section/5_conclusion}
{
    \small
    \bibliographystyle{ieeenat_fullname}
    \bibliography{main_ICCV}
}
\renewcommand{\thetable}{\Alph{table}}
\renewcommand{\thefigure}{\Alph{figure}}

\end{document}

%% file: preamble.tex
\def\eg{\emph{e.g.}} 

\def\ie{\emph{i.e.}}

\def\etal{\emph{et al.}}

\newcommand{\x}{\mathbf{x}}

\newcommand{\z}{\mathbf{z}}
\newcommand{\e}{\mathbf{e}}
\newcommand{\w}{\mathbf{w}}

\newcommand{\bTheta}{\boldsymbol{\Theta}}

\DeclareMathOperator*{\argmax}{argmax}

%% file: section/0_abstract.tex
\begin{abstract}

With the increasing attention to pre-trained vision-language models (VLMs), \eg, CLIP, substantial efforts have been devoted to many downstream tasks, especially in test-time adaptation (TTA). However, previous works focus on learning prototypes only in the textual modality
while overlooking the ambiguous semantics in class names. These ambiguities lead to textual prototypes that are insufficient to capture visual concepts, resulting in limited performance. To address this issue, we introduce \textbf{ProtoMM}, a training-free framework that constructs multimodal prototypes to adapt VLMs during the test time. By viewing the prototype as a discrete distribution over the textual descriptions and visual particles, ProtoMM has the ability to combine the multimodal features for comprehensive prototype learning. More importantly, the visual particles are dynamically updated as the testing stream flows. This allows our multimodal prototypes to continually learn from the data, enhancing their generalizability in unseen scenarios. In addition, we quantify the importance of the prototypes and test images by formulating their semantic distance as an optimal transport problem.
Extensive experiments on 15 zero-shot benchmarks demonstrate the effectiveness of our method, achieving a 1.03\% average accuracy improvement over state-of-the-art methods on ImageNet and its variant datasets.
\end{abstract}

%% file: section/1_introduction.tex
\section{Introduction}
\input{figs/motivation}




Recent advances in vision-language models (VLMs) ~\cite{Flamingo, CLIP, align, LMM-R1, YangWYC023, Mimic} have shown remarkable success across a wide range of visual tasks.
These foundation models, \eg, CLIP~\cite{CLIP}, are pre-trained on web-scale image-text pairs, achieving superior capabilities in addressing zero-shot classification~\cite{coop, TPT, prograd,zhu2025robust}. 
One key factor behind these successes is to find the optimal prototypes for each category. For example, CoOp~\cite{coop} and CoCoOp~\cite{cocoop} introduce learnable prompt vectors concatenated to textual prototypes.
Another line of research, including TPT~\cite{TPT} and C-TPT~\cite{C-TPT}, aims to tune class prompts during inference time for each testing sample, a strategy known as test-time adaptation.


Despite the improvements brought by these methods, we empirically find that the ambiguities in class names result in textual prototypes struggling to differentiate the classes.
As illustrated in Figure~\ref{fig:motivation}(a), the two different classes ``sword lily'' and ``blackberry lily'' exhibit high cosine similarity (0.67) in the CLIP space. This similarity arises not only from the common word ``lily'' but also because both flowers belong to the Iridaceae family. 
As a result, relying on the description-based prototypes makes it challenging to classify the images of  ``sword lily'' and ``blackberry lily'' effectively (0.53 vs. 0.55 and 0.56 vs. 0.58).  
Furthermore, consider the class names ``laptop'' and ``desktop computer'' as depicted in Figure~\ref{fig:motivation}(b). Although they do not share any common same word, they still exhibit a high cosine similarity (0.69) because they both refer to different forms of computers. Consequently, test images of ``laptop'' and ``desktop computer'' have comparable similarities to both class names, making differentiation challenging (0.67 vs. 0.63 and 0.65 vs. 0.65). To this end, recent studies aim to construct high-quality prototypes in various ways, including multiple prompts generation~
\cite{liu2023patch}, cache-based tuning~\cite{TDA}, and diverse data augmentation with diffusions~\cite{DiffTPT}. Unfortunately, all these models focus solely on the textual domain, overlooking the information from the visual aspect~\cite{LinYKPR23, multimodal_SS}.

\input{figs/motivation2}

As discussed in previous studies~\cite{multimodal_SS, XingWCZLWZ24, yang2024learning,zhu2024selective,zhu2024enhancing,qiu2025multimodal,zhu2025project,yang2024robust}, multimodal signals, such as vision and language, provide complementary knowledge that enhances the understanding of objects and class concepts. Although these class names are semantically similar, their visual characteristics provide crucial information for reducing ambiguities.
In this paper, we proposed \textbf{ProtoMM}, a training-free multimodal prototype learning framework for adapting VLMs at test time. Moving beyond prototype learning within the single textual domain, ProtoMM aims to construct multimodal prototypes from both the textual descriptions and image stream. Specifically, for each class, we view its prototype as a discrete distribution over the corresponding label descriptions and visual particles. The visual particles collect the label-specific image features and are updated dynamically as the testing stream flows. This enables our multimodal prototype to accumulate more and more side information, improving the prediction performance. As illustrated in Figure~\ref{fig:motivation2}(a), it is difficult to determine whether the test image belongs to the ``desktop computer'' or ``laptop'' class at time $t_1$. In contrast, with the help of our multimodal prototype learning, one can easily distinguish these two classes at time $t_5$. Furthermore, we compare the KL divergence~\cite{KL} and Maximum Mean Discrepancy (MMD)~\cite{MMD} between our multimodal prototypes and ground truth. The results summarized in Figure~\ref{fig:motivation2}(b) show that as time passes, the distribution discrepancy decreases significantly, \eg, KL divergence from 18.7 to 9.5 and MMD from 0.97 to 0.29.\footnote{$t_1$ and $t_2$ denote the $5000$th and $10000$th sample at the test-time scenario on the ImageNet dataset~\cite{ImageNet}, respectively.}


The main contributions are summarized as follows:
\begin{itemize}
    \item We propose a training-free multimodal prototype learning framework to boost the generalization ability of VLMs in the test-time scenario.
    \item We leverage informative visual knowledge to construct multimodal prototypes, and then update the multimodal prototypes by solving an OT problem.
    \item We demonstrate the effectiveness of our method by conducting experiments on 15 datasets, showing consistent and significant improvements over existing baselines.
\end{itemize}

%% file: figs/motivation.tex
\begin{figure}
  \centering
    \includegraphics[width=1\linewidth]{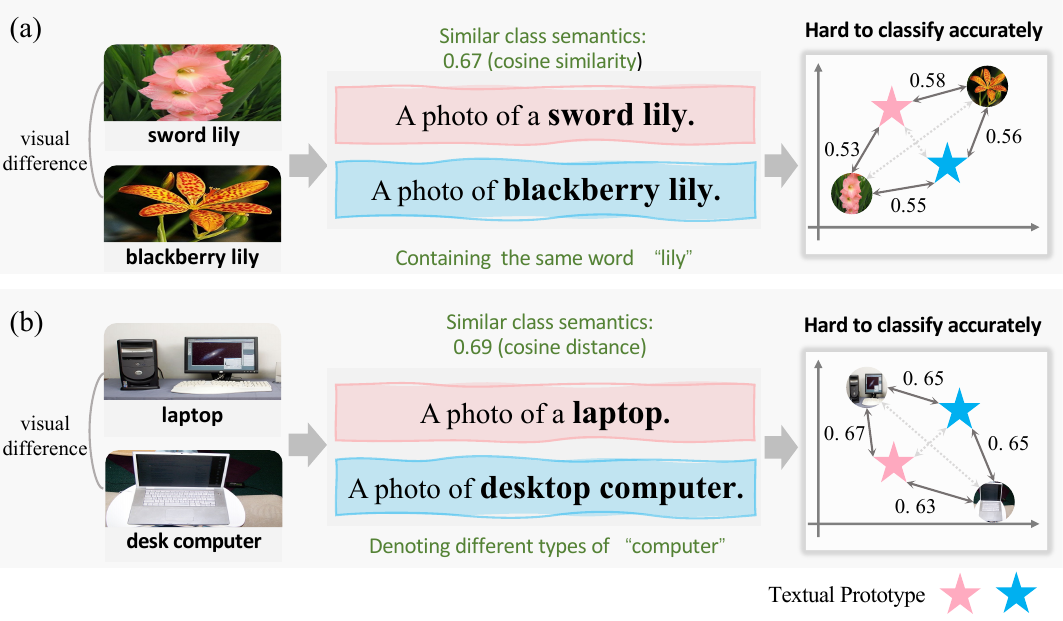}
    \caption{Observations of ambiguities in class names from the Oxford Flowers~\cite{Flower} and ImageNet~\cite{ImageNet}: (a) The class names ``sword lily'' and ``blackberry lily'' both refer to flowers from the Iridaceae family and share the word ``lily''. (b) The class names ``laptop'' and ``desktop computer'' denote different types of computers.}
    \label{fig:motivation}
\end{figure}

%% file: figs/motivation2.tex
\begin{figure}
  \centering
    \includegraphics[width=1\linewidth]{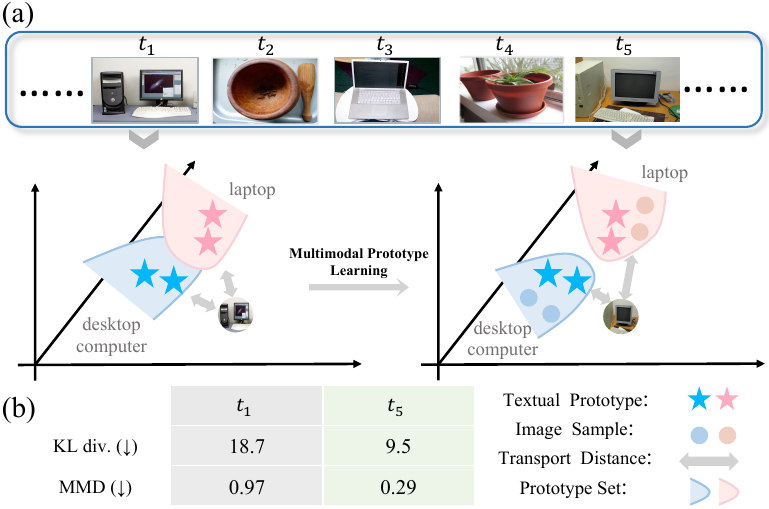}
    \caption{Illustration of multimodal prototype learning on the ImageNet dataset~\cite{ImageNet}. (a) Process of updating multimodal prototypes. (b) Comparisons of distribution metrics.}
    \label{fig:motivation2}
\end{figure}

%% file: section/2_related_work.tex
\section{Related Work}
In this section, we begin with a concise review of test-time adaptation with VLMs, followed by a discussion of optimal transport and its applications in VLMs.

\subsection{Test-time Adaptation}
Test-time adaptation~\cite{TPT, C-TPT, DiffTPT} is an effective technique to address the distribution shift between testing and training datasets. Its core idea is to adjust the model using a streaming of unlabeled test samples. Early works~\cite{Tent} improve test-time adaptation performance by fine-tuning normalization layers. Motivated by the success of VLMs~\cite{CLIP, align}, recent studies~\cite{TPT, C-TPT} propose to optimize textual prompts via entropy minimization. While these methods require high computational costs due to the gradient backpropagation. To address this,  another branch of works~\cite{TDA, dota, DMN}
adapting the VLMs in a non-parametric manner has emerged. Karmanov~\etal~\cite{TDA} designs positive and negative caches to store high-confidence samples, improving the performance by calculating the similarity between the test sample and cached samples. Han~\etal~\cite{dota} proposes a method named DOTA~\cite{dota}, estimating the distribution of test samples progressively. Zhang~\etal~\cite{DMN} extends the cache model to a dual memory network (DMN), which also works in test-time scenarios. Although these methods avoid gradient updates, they all need logit fusion for prediction, and their performance is sensitive to fusion hyperparameters.

Our work aligns with the training-free paradigm. Unlike existing methods that primarily rely on unimodal information (class names), we demonstrate that incorporating multimodal information (test images and class names) facilitates the adaptation of VLMs more effectively. Moreover, we propose a dynamic multimodal prototype learning framework that utilizes visual knowledge from test samples to update the multimodal prototypes. 

\subsection{Optimal Transport}
Optimal transport (OT)~\cite{monge1781memoire} is initially introduced to solve the problem of how to reduce the cost when moving several items simultaneously, and then it is applied to measure the distance between distributions in the deep learning areas~\cite{iLPC, ChenG0LC020, GuoT0ZZ22, PLOT, AWT, li25}, including domain adaption~\cite{CourtyFTR17}, clustering\cite{eWWL17, ShiSY24}, image classification~\cite{ease, GaoZLG23, wang2023tuning, li2023patchct}. 
For example, Lazarou~\etal~\cite{iLPC} directly aligns the distribution of two sets of feature embeddings to address the transductive few-shot task. Guo~\etal~\cite{GuoT0ZZ22} proposes hierarchical OT to transfer the statistics from the base classes to novel classes in distribution calibration. Chen~\etal~\cite{PLOT} aims to align the features of vision and language modalities by optimizing textual prompts via OT. Recently, AWT~\cite{AWT}, a simple and effective method adapts VLMs in zero/few-shot scenarios, which is closely related to our work. AWT formulates the image-text distance calculation as an OT problem, and solving this problem reveals the semantic correlations in the vision-language space for prediction. 

While AWT and the above methods focus on aligning different sets or modalities, our approach leverages OT to enhance textual prototypes by incorporating visual knowledge, rather than aligning different sets or modalities. Specifically, we aim to accumulate informative visual knowledge into textual prototypes to reduce ambiguities. To achieve this, we utilize the transport plan to adjust each test sample's contribution to different class prototypes, and then use the weighted test sample to update the prototypes.





%% file: section/3_method.tex
\section{Method}
In this section, we first briefly review the preliminary knowledge of VLMs and OT~\cite{monge1781memoire}. Then we introduce the details of our multimodal prototype learning framework for adapting VLMs in test-time scenarios as illustrated in Figure~\ref{fig:framework}.

\subsection{Preliminaries}
\noindent{\textbf{Vision-language models (VLMs).}} CLIP~\cite{CLIP} as one of the well-known VLMs, consists of a vision encoder $f(\cdot)$ and a text encoder $g(\cdot)$. Given a set of $C$ class names $\{z_c\}_{c=1}^C$, CLIP first constructs the class description using a pre-defined prompt template $\Phi$, \textit{e.g.}, $\Phi(z_c)$ can be chosen as ``a photo of a $z_c$'' and then obtains the prototypes by feeding the class descriptions into $g$: $\z_c = g(\Phi(z_c))$. For the testing image $x_t$ at time $t$, let $\x_t=f(x_t)$ denotes the extracted visual feature. The zero-shot classification for $x_t$ is computed as:
\begin{equation}\label{eq:fc}
    y_t =\argmax_{c\in\{1,2,\dots,C\}} \z_c^\top\x_t,
\end{equation}
where $\z,\x \in \mathbb{R}^d$ share the same semantic embedding space. $\z_c$ can be viewed as the prototype of $c$-th class. To improve the quality of $\z_c$, previous studies focus on test prompt tuning of $\Phi$ or generating detailed class descriptions via large language models (LLMs)~\cite{metaprompt, CuPL}. 

\vspace{5pt}
\noindent{\textbf{Optimal transport.}} Optimal Transport (OT)~\cite{monge1781memoire} is an efficient tool to measure the distance between two distributions.  
Specifically, consider two discrete distributions in the feature space: 
$P = \sum_{i=1}^{|V|}{a}_i \delta_{\mathbf{v}_i}$ and $Q = \sum_{j=1}^{|U|}{b}_j  \delta_{\mathbf{u}_j}$, where
$\delta$ is the Dirac delta function and $|V|$ and $|U|$ denote the number of points in $P$ and $Q$, respectively. Here, $\mathbf{a}=[a_1,\dots,a_{|V|}]^\top$ and $\mathbf{b}=[b_1,\dots,b_{|U|}]^\top$ denote the probability vectors that sum to 1. Given the cost matrix $\mathbf{C} \in \mathbb{R}^{{|V|} \times {|U|}}$, whose each element measures the transport cost between $\mathbf{v}_i$ and $\mathbf{u}_j$, the OT distance between distributions $P$ and $Q$ is formulated as:
\begin{equation}
\begin{gathered} \label{eq:optimization}
d_{\mathrm{OT}}(P,Q;\mathbf{C}) = \underset{\mathbf{T}}{{\min}}\langle \mathbf{T}, \mathbf{C} \rangle, \\
 \mathrm{s.t.} \ \ \mathbf{T}\mathbbm{1}_{|U|} = \mathbf{a}, \; \mathbf{T}^\top \mathbbm{1}_{|V|} = \mathbf{b},
\end{gathered}
\end{equation}
where $\mathbf{T} \in \mathbb{R}^{|V| \times |U|}$ is the transport plan to be learned, which indicates the transport probability from source point in $P$ to target point in $Q$.
$\langle \cdot, \cdot \rangle$
denotes the Frobenius dot-product and $\mathbbm{1}_{|V|}$ refers to $|V|$-dimensional all-one vector. However, directly computing Eq.~\eqref{eq:optimization} is often computationally intensive. Following previous work~\cite{iLPC,PLOT}, we employ the Sinkhorn algorithm~\cite{Cuturi13}, which introduces an entropic regularization term to facilitate efficient optimization. The resulting objective is then given by:
\begin{equation}
\label{eq:opt_Sink}
d_{\mathrm{OT}}(P,Q;\mathbf{C}) = \underset{\mathbf{T}}{{\min}}\langle \mathbf{T}, \mathbf{C} \rangle - \epsilon h (\mathbf{T}),
\end{equation}
where $h(\cdot)$ is entropy and $\epsilon \geq 0$ is a coefficient.



\subsection{Distributed Feature Construction}

Moving beyond prototype learning within the textual domain, we introduce ProtoMM and explore the multimodal prototype discovery for zero-shot classification. As shown in Figure~\ref{fig:framework}, ProtoMM consists of two main modules: distributed feature construction that models the multimodal prototype and testing image as discrete distributions, and multimodal prototype learning that updates multimodal prototypes by integrating visual information from history.

\input{figs/framework}
To mimic how humans learn a new concept through different perspectives and obtain informative features, we follow strategies from previous works~\cite{AWT, WAC, VisDesc} and view the image and prototype as discrete distribution at the CLIP~\cite{CLIP} space. Specifically, for each image $x_t$, we employ common augmentation techniques such as random crop, flip, and resize. This augments $x_t$ into multiple views $\{x_t^n\}_{n=1}^{N}$, including $N-1$ augmentations and the original one. 
Simultaneously, we query large language models~\cite{metaprompt, CuPL} to generate descriptive sentences for each class name. As a result, each textual prototype $z_c$ is enhanced with $M-1$ additional descriptions, denoted as $\{z_c^m\}_{m=1}^{M}$. Mathematically, the resulting distributions of the image and prototype are formulated as:
\begin{equation}\label{eq:ot}
    P_t=\sum_{n=1}^{N}{a}_t^n \delta_{\x_t^n}; \quad Q_c=\sum_{m=1}^{M}{b}_c^m \delta_{\z_c^m},
\end{equation}
where $\x_t^n$ is the feature of the $n$-th image view and it contributes detailed information about a specific region or angle. $\z_c^m$ is the feature of the $m$-th prototype description from LLMs and it shares reasonable prior knowledge of class $z_c$. Instead of viewing the image and prototype as a single point in the feature space, Eq.~\eqref{eq:ot} models them as discrete distributions in the feature space. This enables our ProtoMM to capture diverse details, leading to comprehensive representation learning. $\mathbf{a}_t =[a_t^1,\cdots,a_t^N]^{\top}$ and $\mathbf{b}_c=[b_c^1,\cdots,b_c^M]^{\top}$ denote the important weights of image and feature points, respectively, \ie, high-quality enhancements should be highly valued while uninformative enhancements should be ignored, which will be discussed later.

As mentioned before, the prototype in Eq.~\eqref{eq:ot} only considers the textual domain, which may not fully cover the visual details of class $z_c$, thereby we introduce the concept of multimodal prototype learning and aim to integrate the visual modality into $Q_c$. Specifically, we expand the prototype set $\{\z_c^m\}_{m=1}^{M}$ by appending $S$ additional visual particles:$\{\e_c^s\}_{s=1}^{S}$, where $\e_c\in\mathbb{R}^d$. Then the multimodal prototype is formulated as (We still use $Q_c$ for convenience):
\begin{equation}\label{eq:mmprototype}
    {Q}_c=\sum_{m=1}^{M}w_c^m \delta_{\z_c^m} + \sum_{s=1}^S w_c^{M+s}\delta_{\e_c^s},
\end{equation}
where $\mathbf{w}_c = [w_c^1,\cdots,w_c^{M+S}]^\top$ denotes the important weights of $M+S$ points in the multimodal prototype. The first term comes from the knowledge of LLMs and the second term can be viewed as a cache that aims to complete the missing features of LLMs using historical images (Sec.~\ref{mpl}). Combining these two modalities, Eq.~\eqref{eq:mmprototype} are expected to cover comprehensive characteristics of the $c$-th class. 

Following previous works~\cite{TDA, dota}, we evaluate test samples sequentially in a streaming manner and hope the visual cache learns from the history, providing valuable guidance for the following samples. Before testing, the visual cache and its importance weights must be properly initialized to complete the multimodal prototype in Eq.~\eqref{eq:mmprototype}. Since images are unavailable at the start of testing, it is natural to construct all visual cache of prototype $P_c$ by averaging corresponding textual descriptions: $\e_c^s = \frac{1}{M}\sum_{m=1}^{M}\z_c^m.$ To determine the importance weights of each element in $P$ and $Q$, we propose a novel calculating strategy based on the input image. For the arriving image $x_t$, we first calculate the important weights of its $N$ visual augments according to all prototypes:
\begin{equation}
\begin{aligned}\label{eq:a_t}
 a_t^n &= \frac{\exp \left(h({\x_t^n})\right)}{\sum_{n^{\prime}=1}^{N} \exp \left(h({\x_t^{n^{\prime}}})\right)}, \\
 h({\x_t^n}) &= \sum_{c=1}^{C}p(\bar{\z}_c|{\x}_t^n){\log}p(\bar{\z}_c|{\x}_t^n),
 \end{aligned}
\end{equation}
where $\bar{\z}_c =\frac{1}{M+S}(\sum_{m=1}^{M}\z_c^m + \sum_{s=1}^{S}\e_c^s)$ denotes the mean of $c$-th multimodal prototype. $a_t^n$ becomes larger when the content of $n$-th augment is semantically close to descriptions at the embedding space and the noise augments such as backgrounds will be ignored by a lower attention value. 

Similarly, the important weight $\w_c$ for $c$-th class is  derived by considering its correlation with the visual features across all augmentations:
\begin{equation}
\begin{aligned}\label{eq:w_c}
    w_c^m &= \frac{\exp \left(h({\z}_c^m)\right)}{\sum_{m^{\prime}=1}^{M+S} \exp \left(h({{\z}_{c}^{m^{\prime}}})\right)},\\
    h({{\z}_c^m}) &= \sum_{n=1}^{N}p({\z}_c^m|{\x}_t^n){\log}p({\z}_c^m|{\x}_t^n),
\end{aligned}
\end{equation}
$\{\z_c^{m}\}_{m={M+1}}^{M+S} = \{\e_c^{s}\}_{s={1}}^{S}$. Intuitively, we use the visual content to measure the importance of each point in $Q$ and expect to construct an instance-specific prototype, improving the generalizability of unseen scenarios.

\subsection{Multimodal Prototype Learning}\label{mpl}
Once $\mathbf{a}_t$ and $\w_c$ are calculated, we construct the discrete distributions $P_t$ and $Q_c$ according to Eq.~(\ref{eq:ot}-~\ref{eq:mmprototype}). Specifying the cost matrix $\Cmat_{tc} \in \mathbb{R}^{N \times(M+S)}$ as the cosine distance between the visual augmentations and multimodal prototypes, we reformulate Eq.~\eqref{eq:opt_Sink} as:
\begin{equation}
\begin{gathered} \label{eq:ot_2}
d_{\mathrm{OT}}(P_t,Q_c;\mathbf{C}_{tc}) = \underset{\mathbf{T}_{tc}}{{\min}}\langle \mathbf{T}_{tc}, \mathbf{C}_{tc} \rangle - \epsilon h (\mathbf{T}_{tc}), \\
 \mathrm{s.t.} \ \ \mathbf{T}_{tc}\mathbbm{1}_{M+S} = \mathbf{a}_t, \; \mathbf{T}_{tc}^\top \mathbbm{1}_{N} = \w_c,
\end{gathered}
\end{equation}
where $\mathbf{T}_{tc}\in\mathbb{R}^{N\times(M+S)}$, whose each element denotes the transport probability from $n$-th augmentation to $m$-th prototype point in the feature space. After solving Eq.~\eqref{eq:ot_2} using the Sinkhorn algorithm~\cite{Cuturi13}, we calculate the prediction probability of image $x_t$ as:
\begin{equation}\label{eq:pred}
    p(y_t=c|\x_t) = \frac{\text{exp}(-d_{\mathrm{OT}}(P_t, Q_c;\mathbf{C}_{tc}))}{\sum_{c'=1}^C \text{exp}(-d_{\mathrm{OT}}(P_t, Q_{c'};\mathbf{C}_{tc'}))}.
\end{equation}

The core idea of the visual cache is to collect high-quality samples for multimodal prototype learning. To this end, we first filter out samples whose highest prediction probability is less than a threshold: $p(y_t=c|\x_t)\geq  \tau$. For the high-quality samples, we develop a score function and only choose the augmentations with high importance. Recalling that the transport plan measures the semantic similarity between the textual prototype and visual augmentations, the score of the augmentation of $x_t$ thus can be calculated as $\bTheta_t = \Tmat_{tc} \wv_{y_t}\in\mathbb{R}^N$. We then select the top-$S$ candidates  $[\x_t^{(1)},\cdots,\x_t^{(S)}]^{\top} \in \mathbb{R}^{S \times d}$ according to the decreasing value sequence $ [{{\theta}}_t^{(1)},\cdots,\theta_t^{(S)}]$ in $\bTheta_t$. The visual cache in our multimodal prototype $Q_c$ is updated using these selected high-quality candidates:
\begin{equation}\label{update}
\begin{gathered}
    \mathbf{e}_c^s \leftarrow \frac{w_t^{M+s} \mathbf{e}_c^s + \theta_t^{(s)} \mathbf{x}_t^{(s)}}
    {w_t^{M+s} + \theta_t^{(s)}},  \\
\end{gathered}
\ \ s \in [1,S], \ c \in [1,C],
\end{equation}
This weighting mechanism ensures that the model can dynamically adjust its focus based on the reliability and relevance of each sample.

%% file: figs/framework.tex
\begin{figure*}
  \centering
    \includegraphics[width=1\linewidth]{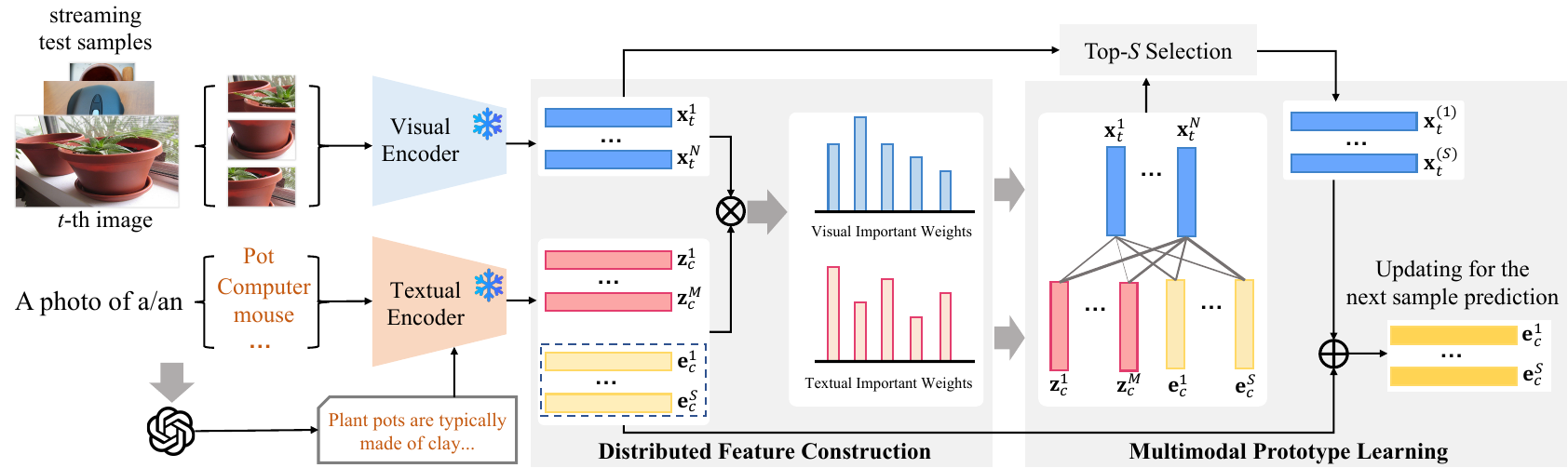}
    \caption{An framework of the proposed method (\textbf{ProtoMM}), which consists of two modules, \ie, (a) \textbf{Distributed Feature Construction}: Expand the textual prototypes with visual features from testing samples. (b) \textbf{Multimodal Prototype Learning}: Updating the multimodal prototypes through the transport plan for the next prediction.}
    \label{fig:framework}
\end{figure*}

%% file: section/4_experiments.tex
\begin{algorithm}[tb]
\footnotesize
   \caption{Pipeline of \textbf{ProtoMM}}
   \label{alg:logic}
\begin{algorithmic}[1]
   \REQUIRE 
   Augmented text prototypes $\{\z_c^m\}_{m=1,c=1}^{M,C}$.   \\
   \FOR{$t = 1,2,\dots,$}
       \STATE Augment the test image $x_t$, and then compute the image features $\{
       \x_t^n\}_{n=1}^{N}$.
       \STATE Construct multimodal features by Eq.~\eqref{eq:mmprototype}.
       \STATE Calculate importance weights $\mathbf{a}_t$ and $\w_c$ using Eq.~\eqref{eq:a_t} and Eq.~\eqref{eq:w_c}, respectively.
       \STATE Obtain transport plan $\mathbf{T}_{tc}$ by solving Eq.~\eqref{eq:ot_2}.
       \STATE Calculate predicted label $p(y_t=c|\x_t)$ by Eq.~\eqref{eq:pred}.
       \STATE Update multimodal prototypes $\{\e_c^s\}_{s=1,c=1}^{S,C}$ via Eq.~\eqref{update}.
   \ENDFOR
\end{algorithmic}
\end{algorithm}

\section{Experiments}
We present the experimental results of our method on 15 classification datasets in the test-time scenario, including performance comparisons, ablation studies, and visualizations. More results are provided in the Supplementary Material.

\subsection{Setup}
\noindent{\textbf{Datesets.}} Following the previous works~\cite{TDA, WAC}, we cover 11 classification datasets including diverse range categories, generic objects (ImageNet~\cite{ImageNet}, Caltech~\cite{Caltech}), scenes (SUN~\cite{SUN}), textures (DTD~\cite{DTD}), satellite images (EuroSAT~\cite{Eurosat}), actions (UCF~\cite{UCF101}) and fine-grained categories (Pets~\cite{OxfordPet}, Cars~\cite{Cars}, Flowers~\cite{Flower}, Food~\cite{Food-101}, Aircraft~\cite{FGVC}).
Additionally, we evaluate on 4 ImageNet distribution shifted datasets~\cite{ImageNet}: ImageNetV2~\cite{ImageNetV2}, ImageNet-Sketch~\cite{ImageNetSketch}, ImageNet-A~\cite{ImageNetA}, ImageNet-R~\cite{ImageNetR}.

\noindent{\textbf{Implementation details.}} The experiments are built upon the pre-trained CLIP~\cite{CLIP} model. We adapt ResNet50 (RN50) and ViT-B/16 backbones as the visual encoder, and the textual encoder is a Transformer~\cite{transformer}. We adhere to the same image augmentation strategies as AWT~\cite{AWT}, including random cropping and resizing as well as flipping. For textual augmentation, we query GPT-3.5 to generate the class descriptions following the previous works~\cite{metaprompt, AWT}. The total augmentation number $M$ and $N$ are both set as 50. Top-$S$ is set to 25, and threshold $\tau = 0.8$. All experiments are conducted on one NVIDIA RTX 3090 GPU.

\noindent{\textbf{Comparison methods.}} We compare our method with previous state-of-the-art (SOTA) test-time adaptation methods, including TPT~\cite{TPT}, DiffTPT~\cite{DiffTPT}, TDA~\cite{TDA}, DOTA~\cite{dota}, DPE~\cite{DPE} and AWT~\cite{AWT}. In these comparison methods, TDA~\cite{TDA}, AWT~\cite{AWT}, and DOTA~\cite{dota} are designed in a training-free manner. TPT~\cite{TPT} and DiffTPT~\cite{DiffTPT} are prompt tuning methods, and DPE~\cite{DPE} belongs to the adapter tuning method. The optimization process of there methods are happended during the inference stage.
Additionally, we compare several prompt learning methods, such as CoOp~\cite{coop} and CoCoOp~\cite{cocoop}, as well as augmentation-based methods, including CUPL~\cite{CuPL}, VisDescs~\cite{VisDesc}, and MetaPrompting~\cite{metaprompt}.
\input{tables/main_results}
\input{tables/main_results_imagenet}
\subsection{Main Results}
We compare our methods with state-of-the-art (SOTA) methods across 11 widely used classification datasets, using RN50 and ViT-B/16 backbones. The results presented in Table~\ref{tab:cross-domain} highlight the superior performance of our method which consistently outperforms AWT~\cite{AWT} across all datasets under RN50 excluding Pets, with an averaged accuracy gain of  0.93\%. When compared to TDA~\cite{TDA} and DOTA~\cite{dota} using RN50, our method achieves the 2.04\% and 0.94\%  averaged accuracy improvements, which is significant in test-time scenarios. In the Caltech101~\cite{Caltech} dataset, DPE~\cite{DPE} exhibits the highest performance, which is due to tuning the adapters for each testing sample. However, this process is more computationally and time-costing. For the ViT-B/16 backbone, we observe similar results, with our method surpassing the existing methods across most datasets, showing an average accuracy improvement of 1.08\% over DPE. Especially on the fine-grained Flowers102~\cite{Flower} datasaet, our method outperforms the second-best AWT by over 2.3\%.
We also evaluate our method with SOTA methods on the ImageNet~\cite{ImageNet} and its 4 out-of-distribution (OOD) variants, as shown in Table~\ref{tab:ood-main}. The experimental results demonstrate that our method exhibits competitive generalization performance across all methods. Specifically, our ProtoMM outperforms TDA by 2.58\% and AWT by 0.71\% on average when using the RN50 backbone. Moreover, with the ViT-B/16 backbone, our method surpasses the second-best method, DOTA, by an average of 1.17\% on the OOD dataset. Even compared to the training-based DPE, it achieves an average improvement of 1.03\%. These results demonstrate the adaptability of our method to varied distributions. Notably, this generalization is achieved without any test-time optimization or tuning.


\subsection{Ablation Study}
We conduct experiments to assess the effectiveness of each module and strategy designed in our method. Additionally, we evaluate its efficiency and present visualizations.

\vspace{5pt}
\noindent{\textbf{Effectiveness of using multimodal prototypes.}} We examine the contributions of the distributed feature construction module and the multimodal prototype learning module, respectively. The results are summarized in Table~\ref{tab:diff_modules}. Row (1) presents the zero-shot CLIP~\cite{CLIP} performance as a baseline. In Row (2), Eq.~\eqref{eq:mmprototype} stands for the distributed feature construction,  which expands the textual prototypes by incorporating additional features with their importance weights, achieving 1\% gains over CLIP on ImageNet with RN50. Row (3) shows the results of using both modules in our method, outperforming the variant that only uses Eq.~\eqref{eq:mmprototype} by over 1\% on Caltech101 with ViT-B/16. These improvements indicate that dynamically incorporating multimodal prototypes leads to better representations and improved test-time adaptation.

\input{tables/abltion_module}
\vspace{5pt}
\noindent{\textbf{Effectiveness of the Number of Augmentations.}}
We evaluate the impact of the number of augmentations on images and class names for both RN50 and ViT-B/16 backbones, as illustrated in Figure~\ref{fig:img_aug} and Figure~\ref{fig:txt_aug}, respectively. For visual augmentations, the accuracy initially improves as the number of augmentations increases from 1 to 40, after which the performance stabilizes. This trend indicates that increasing the number of image augmentations can effectively improve the discriminability of multimodal prototypes up to a certain point.
On the other hand, textual augmentations (Figure~\ref{fig:txt_aug}) show similar results to visual augmentations. As the number of textual augmentations varies from 1 to 70, the accuracy initially increases and then converges. This implies that beyond a certain number, adding more descriptive sentences for class names does not improve the quality of prototypes.
Based on these observations, we set the number of visual and textual augmentations both as $L=K=50$, which is consistent with the setting in AWT.
\input{figs/augmentation}
\input{figs/threshold_topc}

\vspace{5pt}
\noindent{\textbf{{Effectiveness of threshold $\tau$}}}
We set the threshold to ensure confident prediction for the test samples. A large value of $\tau$ implies stricter criteria for selecting samples, which can increase the reliability of predictions. We vary $\tau$ from 0.1 to 0.99. The results in Figure~\ref{fig:threshold}  show that as the threshold $\tau$ increases, performance improves gradually. However, once the threshold exceeds 0.8, the performance begins to degrade. Without testing samples ($\tau$ = 1) fail to establish the multimodal prototypes, leading to worse performance. Furthermore, using all test samples ($\tau$ = 0) without any confidence-based filtering does not lead to optimal performance, as low-confidence predictions may introduce incorrect labels that provide limited useful information and potentially disrupt the structure of the multimodal prototypes.

\vspace{5pt}
\noindent{\textbf{{Effectiveness of top-$S$ selection.}}}
In our method, we select the top-$S$ confident samples for constructing multimodal prototypes. To assess the influence of the number of selected samples, we vary the selection number from 5 to 50 and analyze the results illustrated in Figure~\ref{fig:topc}. The results indicate that performance improves as the number of selected samples increases, reaching a stable state when exceeding a certain number. Including too many augmented samples with high prediction confidence does not further boost performance, as the visual knowledge provided by each test sample to the textual prototypes is sufficient. Overall, our method shows robust performance across a typical range of top-$S$ values, maintaining stable classification accuracy.

\vspace{5pt}
\noindent{\textbf{{Visualization of transport plan.}}}
To intuitively demonstrate the effectiveness of our method, we visualize the heatmap of the transport plan $\mathbf{T}$ for the testing sample, as illustrated in Figure~\ref{fig:visualization}. We select four test images from the ImageNet dataset, such as “hamster,” “mountain bike,” “robin,” and “malamute,” where the transport plans primarily focus on the objects. This demonstrates that our method effectively transfers discriminative visual knowledge to textual prototypes, resulting in more representative multimodal prototypes.
\input{figs/heatmap}

\vspace{5pt}
\noindent{\textbf{{Comparison of Inference Time}}}
Our method is training-free, unlike methods that require prompt tuning or adapter tuning, such as TPT, DiffTPT, and DPE, which involve backpropagating through an expensive encoder or adapter during optimization. We evaluate the inference time using a single NVIDIA 3090 GPU. The experimental results in Table~\ref{tab:time} indicate that our method is faster than the tuning-based methods. Although our method takes slightly more time than TDA, we achieve a notable performance improvement, outperforming TDA by over 2.4\%. It is worth noting that the Sinkhorn algorithm~\cite{Cuturi13} supports efficient parallel computation, allowing the distance between one distribution and multiple others to be evaluated simultaneously, as demonstrated in prior works~\cite{PLOT, AWT}. This parallelism further contributes to the comparable runtime of our method relative to TDA.
\input{tables/time_consume}




%% file: tables/main_results.tex
\begin{table*}[ht]
  \centering

  \resizebox{0.85\linewidth}{!}{
    \begin{tabular}{ll*{12}{c}}
      \toprule
      \multicolumn{1}{l}{} &Method & \rotatebox{90}{Aircraft} & \rotatebox{90}{Caltech101} & \rotatebox{90}{Cars} & \rotatebox{90}{DTD} & \rotatebox{90}{EuroSAT} & \rotatebox{90}{Flower102} & \rotatebox{90}{Food101} & \rotatebox{90}{Pets} & \rotatebox{90}{SUN397} & \rotatebox{90}{UCF101} & \rotatebox{90}{ImageNet} & \rotatebox{90}{Average} \\
      \midrule
      \multicolumn{1}{l|}{} & CLIP~\cite{CLIP} & 16.11 & 87.26 & 55.89 & 40.37 & 25.79 & 62.77 & 74.82 & 82.97 & 60.85 & 59.48 & 59.81 & 56.92  \\
      \cmidrule{2-14}
      \multicolumn{1}{l|}{} & CoOp~\cite{coop} & 15.12 & 86.53 & 55.32 & 37.29 & 26.20 & 61.55 & 75.59 & 87.00 & 58.15 & 59.05 & 63.33 & 56.83 \\
      \multicolumn{1}{l|}{} & CoCoOp~\cite{cocoop} & 14.61 & 87.38 & 56.22 & 38.53 & 28.73 & 65.57 & 76.20 & 88.39 & 59.61 & 57.10 &  62.81 & 57.74  \\
      \cmidrule{2-14}
      \multicolumn{1}{l|}{} & CuPL~\cite{CuPL} &19.59 &89.29 &57.28 &48.64 &38.38 &65.44 &76.94 & 84.84 &62.55 &58.97 &61.45 & 60.31\\
      \multicolumn{1}{l|}{} & VisDesc~\cite{VisDesc} &16.26 &88.11 &54.76 &41.96 &37.60 &65.37 &76.80 &82.39 &59.84 &58.47 &59.68 & 58.29 \\
      \cmidrule{2-14}
      \multicolumn{1}{l|}{} & TPT~\cite{TPT} & 17.58 & 87.02 & 58.46 & 40.84 & 28.33 & 62.69 & 74.88 & 84.49 & 61.46 & 60.82 & 60.74 & 57.93 \\
      \multicolumn{1}{l|}{} & DiffTPT~\cite{DiffTPT} & 17.60 & 86.89 & \textbf{60.71} & 40.72 & 41.04 & 63.53 & \textbf{79.21} & 83.40 & 62.72 & 62.67 & 60.80 & 59.93 \\ \cmidrule{2-14}
      \multicolumn{1}{l|}{} & TDA~\cite{TDA} & 17.61 & 89.70 & 57.78 & 43.74 & 42.11 & \underline{68.74} & 77.75 & 86.18 & 62.53 & 64.18 & 61.35 & 61.06 \\
      \multicolumn{1}{l|}{} & DOTA~\cite{dota} & 18.06 & 88.84 & 58.72 & 45.80 & \textbf{47.15} & 68.53 &\underline{78.61} & 87.33 & 63.89 & \underline{65.08} &61.82 & 62.16\\
      \multicolumn{1}{l|}{} & DPE~\cite{DPE} & 19.80 & \textbf{90.83} & 59.26 & \underline{50.18} & 41.67 & 67.60 & 77.83 & 85.97 & 64.23 & 61.98 &\underline{63.41} & 62.06\\
      \multicolumn{1}{l|}{} & AWT$^*$~\cite{AWT} &\underline{20.31}  &89.57  &59.22  &49.46  &38.80  & 68.21  &77.18  &\textbf{88.22}  &\underline{64.91}  &65.02  &63.01 & \underline{62.17} \\ \cmidrule{2-14}
      \multicolumn{1}{l|}{\multirow{-12}{*}{\rotatebox{90}{RN50}}} & \textbf{ProtoMM} &\textbf{21.21} &\underline{89.94} &\underline{60.10} &\textbf{50.53} &\underline{42.35} &\textbf{69.95} &77.31 &\underline{87.92} &\textbf{65.74} &\textbf{65.34} &\textbf{63.76} & \textbf{63.10} \\
      \midrule
      \multicolumn{1}{l|}{} & CLIP~\cite{CLIP} & 23.22 & 93.55 & 66.11 & 45.04 & 50.42 & 66.99 & 82.86 & 86.92 & 65.63 & 65.16 & 68.34 & 64.93\\
      \cmidrule{2-14}
      \multicolumn{1}{l|}{} & CoOp~\cite{coop} & 18.47 & 93.70 & 64.51 & 41.92 & 46.39 & 68.71 & 85.30 & 89.14 & 64.15 & 66.55 & 71.51 & 64.57\\
      \multicolumn{1}{l|}{} & CoCoOp~\cite{cocoop} & 22.29 & 93.79 & 64.90 & 45.45 & 39.23 & 70.85 & 83.97 & 90.46 & 66.89 & 68.44 & 71.02 & 65.20 \\
      \cmidrule{2-14}
      \multicolumn{1}{l|}{} & CuPL~\cite{CuPL} &24.90 &92.98 &65.29 &44.56 &47.84 &71.30 &86.11 &89.13 &62.59 &66.83 &69.62 & 65.55\\
      \multicolumn{1}{l|}{} & Visdesc~\cite{VisDesc} &24.30 &94.60 &64.08 &44.98 &54.84 &70.85 &85.05 &88.85 &67.99 &67.12 &68.55 & 66.47\\
      \cmidrule{2-14}
      \multicolumn{1}{l|}{} & TPT~\cite{TPT} & 24.78 & 94.16 & 66.87 & 47.75 & 42.44 & 68.98 & 84.67 & 87.79 & 65.50 & 68.04 & 68.98 & 65.45 \\
      \multicolumn{1}{l|}{} & DiffTPT~\cite{DiffTPT} & 25.60 & 92.49 & 67.01 & 47.00 & 43.13 & 70.10 & \textbf{87.23} & 88.22 & 65.74 & 62.67 & 70.30 & 65.40 \\ \cmidrule{2-14}
      \multicolumn{1}{l|}{} & TDA~\cite{TDA} & 23.91 & 94.24 & 67.28 & 47.40 & \underline{58.00} & 71.42 & 86.14 & 88.63 & 67.62 & 70.66 & 69.51 & 67.71\\
      \multicolumn{1}{l|}{} & DOTA~\cite{dota} & 25.59 & 94.32 & 69.48 & 47.87 & 57.65 & 74.67 & \underline{87.02} & 91.69 & 69.70 & \textbf{72.06} & 70.68 & 69.15 \\
      \multicolumn{1}{l|}{} & DPE~\cite{DPE} & 28.95 & 94.81 & 67.31 & 54.20 & 55.79 & 75.07 & 86.17 & 91.14 & \underline{70.07} & 70.44 &\underline{71.91} & 69.62\\
      \multicolumn{1}{l|}{} & AWT$^*$~\cite{AWT}  & \underline{29.22} & \underline{95.40} & \underline{69.80} & \underline{55.56} & \textbf{58.40} & \underline{75.07} & 85.54 & \textbf{92.23} & 70.00 & 70.70 &71.26 &\underline{70.28}\\ \cmidrule{2-14}
      \multicolumn{1}{l|}{\multirow{-12}{*}{\rotatebox{90}{ViT-B/16}}} & \textbf{ProtoMM} &\textbf{31.02} &\textbf{95.70} &\textbf{69.92} &\textbf{56.38} & 56.11 &\textbf{77.40} &85.89 &\underline{91.90} &\textbf{70.78} &\underline{71.76} &\textbf{72.01} &\textbf{70.70} \\
      \bottomrule
    \end{tabular}
  }
  \caption{{Performance comparisons of  accuracy ($\%$) on the cross-domain benchmark.} All the compared methods are built upon CLIP-RN50 or CLIP-ViT-B/16 backbone. The evaluation metric \textit{Average} is calculated by taking the mean accuracy across all 11 datasets. The best and second best results are denoted in \textbf{bold} and \underline{underline}, respectively. $^*$  refers to our implementation using their publicly released code.}
  \label{tab:cross-domain}
\end{table*}

%% file: tables/main_results_imagenet.tex
\begin{table*}[ht]
  \centering
  \resizebox{0.85\linewidth}{!}{
  \begin{tabular}{l|lccccccc}
    \toprule
    \multicolumn{1}{l|}{} &{Method}      & ImageNet    & ImageNet-A    & ImageNet-V2    & ImageNet-R  & ImageNet-S    & {Average}  &{OODAverage}     \\
  \cmidrule{2-9}
  \multicolumn{1}{l|}{} &CLIP~\cite{CLIP} & 59.81 & 23.24  & 52.91	&{60.72}   &35.48&	46.43&	43.09           \\ 
  \cmidrule{2-9}
  \multicolumn{1}{l|}{} &CoOp~\cite{coop}        & 63.33 & 23.06 & 55.40 & 56.60 & 34.67 & 46.61 & 42.43       \\
  \multicolumn{1}{l|}{} &CoCoOp~\cite{cocoop}    & 62.81 & 23.32 & 55.72 & 57.74 & 34.48 & 46.81 & 42.82   \\
  \cmidrule{2-9}
  \multicolumn{1}{l|}{} &CuPL~\cite{CuPL}          & 61.45 & 23.86 & 54.61 & 61.02 & 35.13 & 47.21 & 43.65    \\
  \multicolumn{1}{l|}{} &VisDes~\cite{VisDesc}  & 59.68 & 21.17 & 52.36 & 57.16 & 33.78 & 44.83 & 41.11 \\
  \cmidrule{2-9}
  \multicolumn{1}{l|}{} &TPT~\cite{TPT}          & 60.74 & 26.67 & 54.70 & 59.11 & 35.09 & 47.26 & 43.89    \\
  \multicolumn{1}{l|}{} &DiffTPT~\cite{DiffTPT}  & 60.80 & \textbf{31.06} & 55.80 & 58.80 & 37.10 & 48.71 & 45.69 \\
  \cmidrule{2-9}
  \multicolumn{1}{l|}{} &TDA~\cite{TDA}          & 61.35 & 30.29 & 55.54 & 62.58 & 38.12 & 49.58 & 46.63 \\
  \multicolumn{1}{l|}{} &DOTA~\cite{dota}         & 61.82 & 30.81 & 55.27 & 62.81 & 37.52 & 49.64 & 46.60 \\
  \multicolumn{1}{l|}{} &DPE~\cite{DPE}          & \underline{63.41} & 30.15 &\underline{56.72} & \underline{63.72} & \underline{40.03} & \underline{50.80}  & \underline{47.65}  \\
  \multicolumn{1}{l|}{} &AWT$^*$~\cite{AWT}     &63.01  &\underline{30.93}  &56.31  &63.06  &39.71  & 50.60 & 47.50 \\
  \cmidrule{2-9}
  \multicolumn{1}{l|}{\multirow{-12}{*}{\rotatebox{90}{RN50}}} &\textbf{ProtoMM}           &\textbf{63.76}  &30.56  &\textbf{57.28}  &\textbf{63.95}  & \textbf{41.05}  & \textbf{51.32} & \textbf{48.21} \\
  \midrule
  \multicolumn{1}{l|}{} &CLIP~\cite{CLIP}        & 68.34 & 49.89 & 61.88 & 77.65 & 48.24 & 61.20 & 59.42             \\
  \cmidrule{2-9}
  \multicolumn{1}{l|}{} &CoOp~\cite{coop}        & 71.51 & 49.71 & 64.20 & 75.21 & 47.99 & 61.72 & 59.28  \\
  \multicolumn{1}{l|}{} &CoCoOp~\cite{cocoop}    & 71.02 & 50.63 & 64.07 & 76.18 & 48.75 & 62.13 & 59.91  \\
  \cmidrule{2-9}
    \multicolumn{1}{l|}{} &CuPL~\cite{CuPL}          & 69.62& 50.72 & 63.27 & 77.05 & 49.02 & 61.93 & 60.01    \\
  \multicolumn{1}{l|}{} &VisDes~\cite{VisDesc}  & 68.55 & 49.07 & 61.80 & 75.13 & 47.97 & 60.50 & 58.49 \\
  \cmidrule{2-9}
  \multicolumn{1}{l|}{} &TPT~\cite{TPT}          & 68.98 & 54.77 & 63.45 & 77.06 & 47.94 & 62.44 & 60.81 \\
  \multicolumn{1}{l|}{} &DiffTPT~\cite{DiffTPT}  & 70.30 & 55.68 & 65.10 & 75.00 & 46.80 & 62.28 & 60.52 \\ \cmidrule{2-9}
  \multicolumn{1}{l|}{} &TDA~\cite{TDA}          & 69.51 & 60.11 & 64.67 & 80.24 & 50.54 & 65.01 & 63.89 \\
  \multicolumn{1}{l|}{} &DOTA~\cite{dota}         & 70.68 & 
  \underline{61.19} & 64.41 & \textbf{81.17} & 51.33 & 65.75 & \underline{64.52} \\
  \multicolumn{1}{l|}{} &DPE~\cite{DPE}          & \underline{71.91} & 59.63 & \underline{65.44} & 80.44 & \textbf{52.26} & \underline{65.93} & 64.44 \\
  \multicolumn{1}{l|}{} &AWT$^*$~\cite{AWT}     & 71.26 & 60.33 & 65.05 & 80.38 & 51.60 & 65.72 & 64.34  \\
  \cmidrule{2-9}
  \multicolumn{1}{l|}{\multirow{-11}{*}{\rotatebox{90}{ViT-B/16}}} &\textbf{ProtoMM}           &\textbf{72.01}  &\textbf{64.02}  &\textbf{65.93}  &\underline{80.87}  &\underline{51.97} & \textbf{66.96} & \textbf{65.69}  \\
    \bottomrule
  \end{tabular}
}
\caption{{Performance comparison of accuracy ($\%$) on the out-of-distribution benchmark}. All the compared methods are built upon CLIP-RN50 or CLIP-ViT-B/16 backbone. The two evaluation metrics \textit{Average} and \textit{OOD Average} are calculated by taking the mean accuracy across all five datasets and four OOD datasets excluding ImageNet. The best and second best results are denoted in \textbf{bold} and \underline{underline}, respectively. $^*$  refers to our implementation using their publicly released code.  
}
\label{tab:ood-main}
\end{table*}

%% file: tables/abltion_module.tex
\begin{table}[!t]
\centering
\tabstyle{3pt}  
\begin{tabular}{lcccccc}
\toprule
& \multirow{2}{*}{Eq.~\eqref{eq:mmprototype}} & \multirow{2}{*}{Eq.~\eqref{update}} & \multicolumn{2}{c}{ImageNet} & \multicolumn{2}{c}{Caltech101} \\ 
\cmidrule(lr){4-5} \cmidrule(lr){6-7}
 &  &  & RN50 & ViT-B/16  & RN50 & ViT-B/16  \\ 
\midrule
(1) & \multicolumn{2}{c}{CLIP~\cite{CLIP}}  & 59.81 & 68.34 & 87.26 & 93.55 \\ 
\midrule
(2) & \ding{52} &  & 60.82  & 68.98   & 88.16  & 94.65 \\  
\midrule
(3) & \textbf{\ding{52}} & \textbf{\ding{52}} & \textbf{63.76} & \textbf{72.01}  & \textbf{89.94} & \textbf{95.70}  \\
\bottomrule
\end{tabular}
\caption{Performance of our method with different modules, where Eq.\eqref{eq:mmprototype} and Eq.\eqref{update} denote the Distributed Feature Construction and Multimodal Prototype Learning in our method, respectively.}
\label{tab:diff_modules}
\end{table}

%% file: figs/augmentation.tex
\begin{figure}[t]
  \centering
  \begin{subfigure}{0.49\linewidth}
  \centering
    \includegraphics[width=1\linewidth]{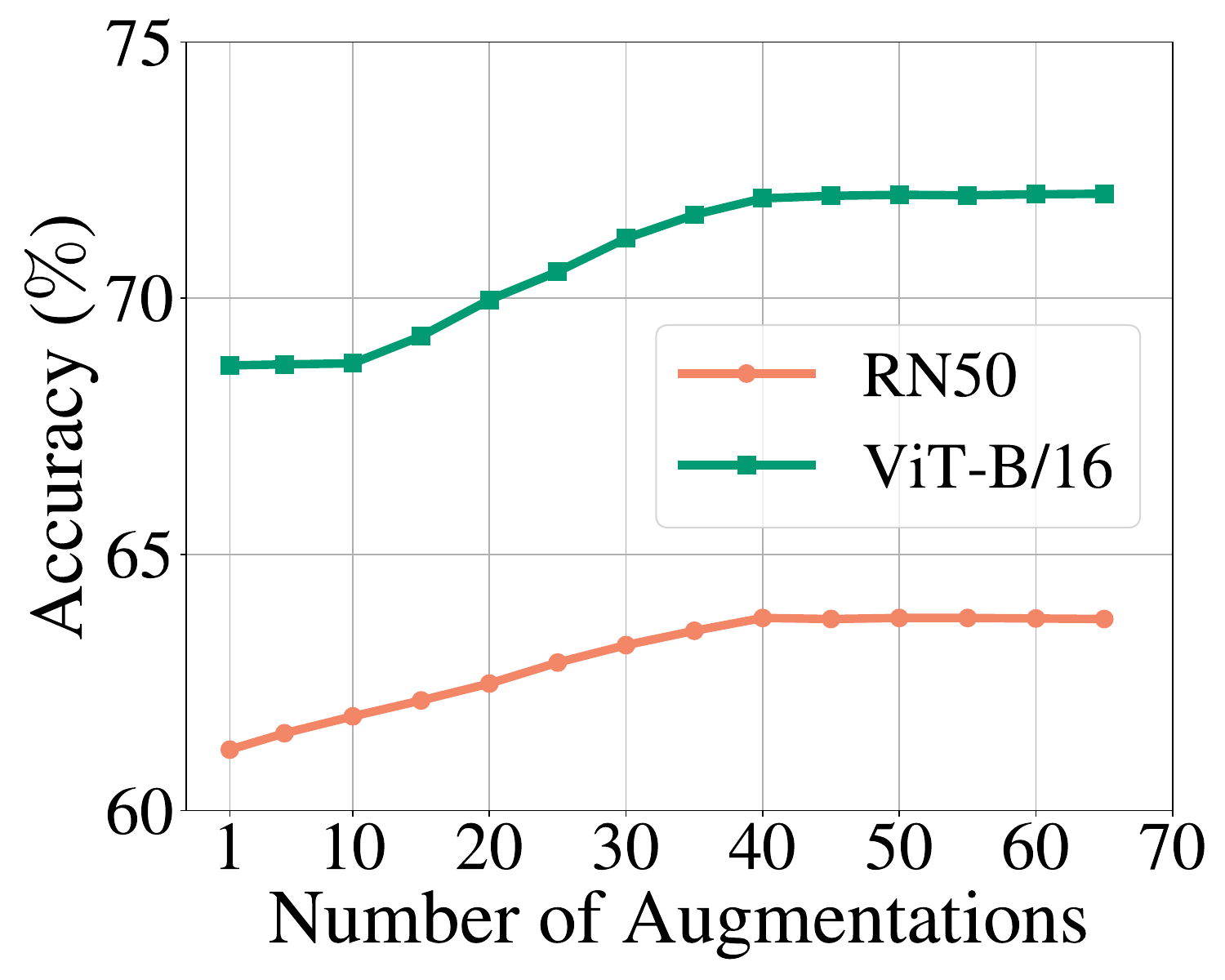}
    \caption{Visual Augmentation.}
    \label{fig:img_aug}
  \end{subfigure}
 \begin{subfigure}{0.49\linewidth}
    \centering
    \includegraphics[width=1\linewidth]{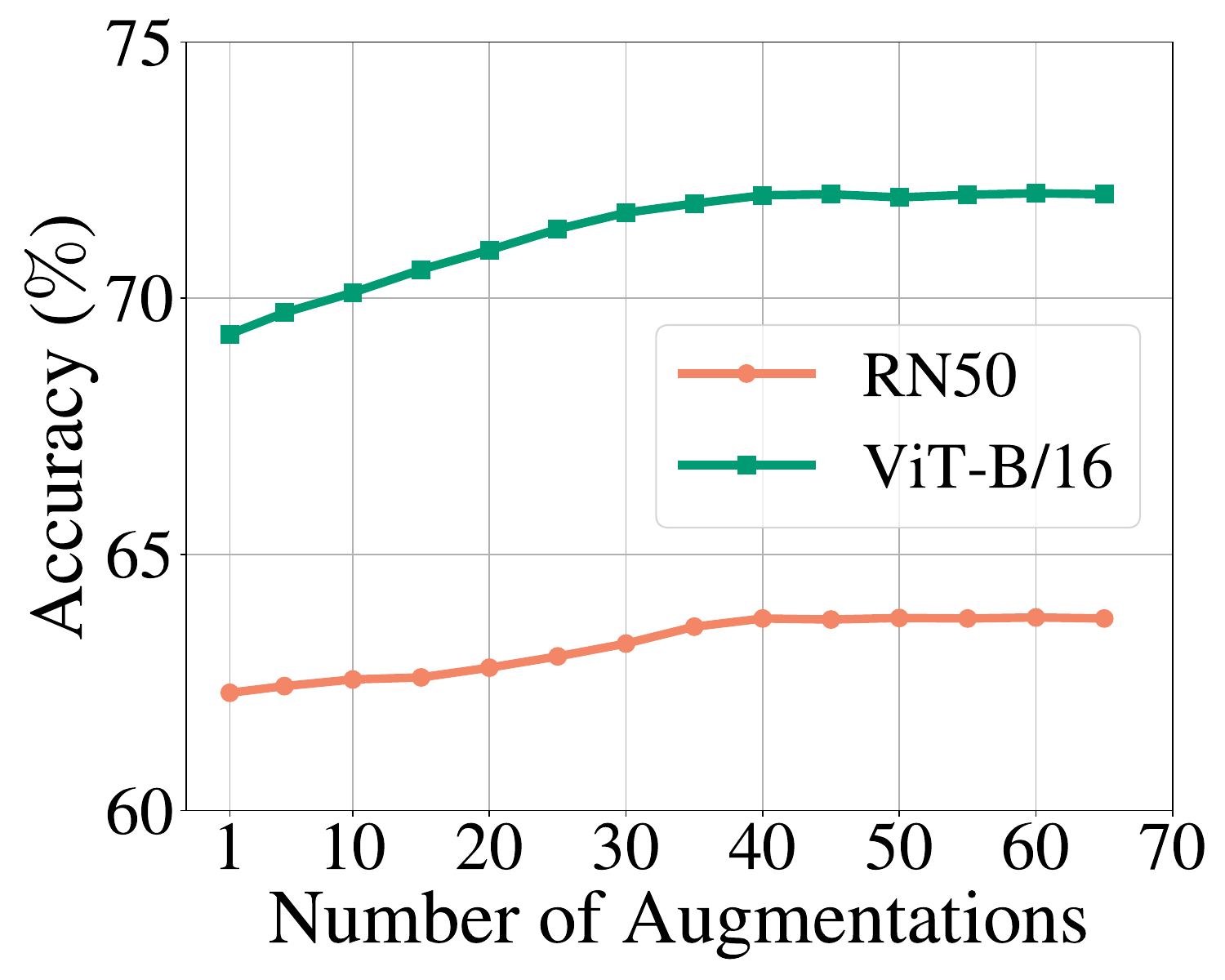}
    \caption{Textual Augmentaion.}
    \label{fig:txt_aug}
  \end{subfigure}
  \caption{Analysis of classification performance by varying the number of augmentations in the ImageNet dataset: (a) image augmentation, (b) class name augmentation.} 
  \vspace{-0.2cm}
  \label{fig:aug}
\end{figure}

%% file: figs/threshold_topc.tex

\begin{figure}[htbp]
\centering
\begin{minipage}{0.46\linewidth}
\centering
\includegraphics[width=1\textwidth]{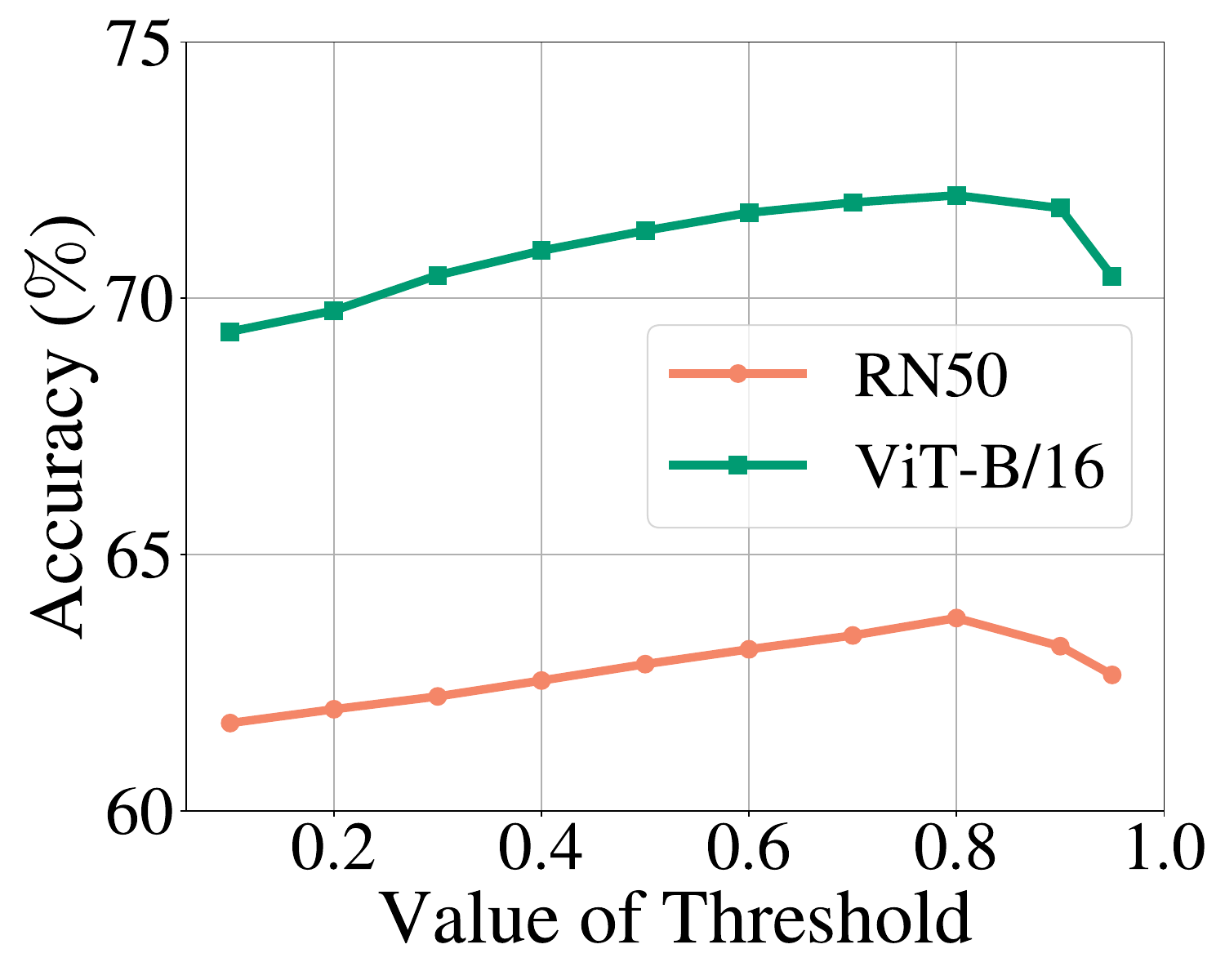}
\caption{Analysis of classification performance by varying the threshold value on the ImageNet dataset.} \label{fig:threshold}
\end{minipage}
\quad
\begin{minipage}{0.46\linewidth}
\centering
\includegraphics[width=1\textwidth]{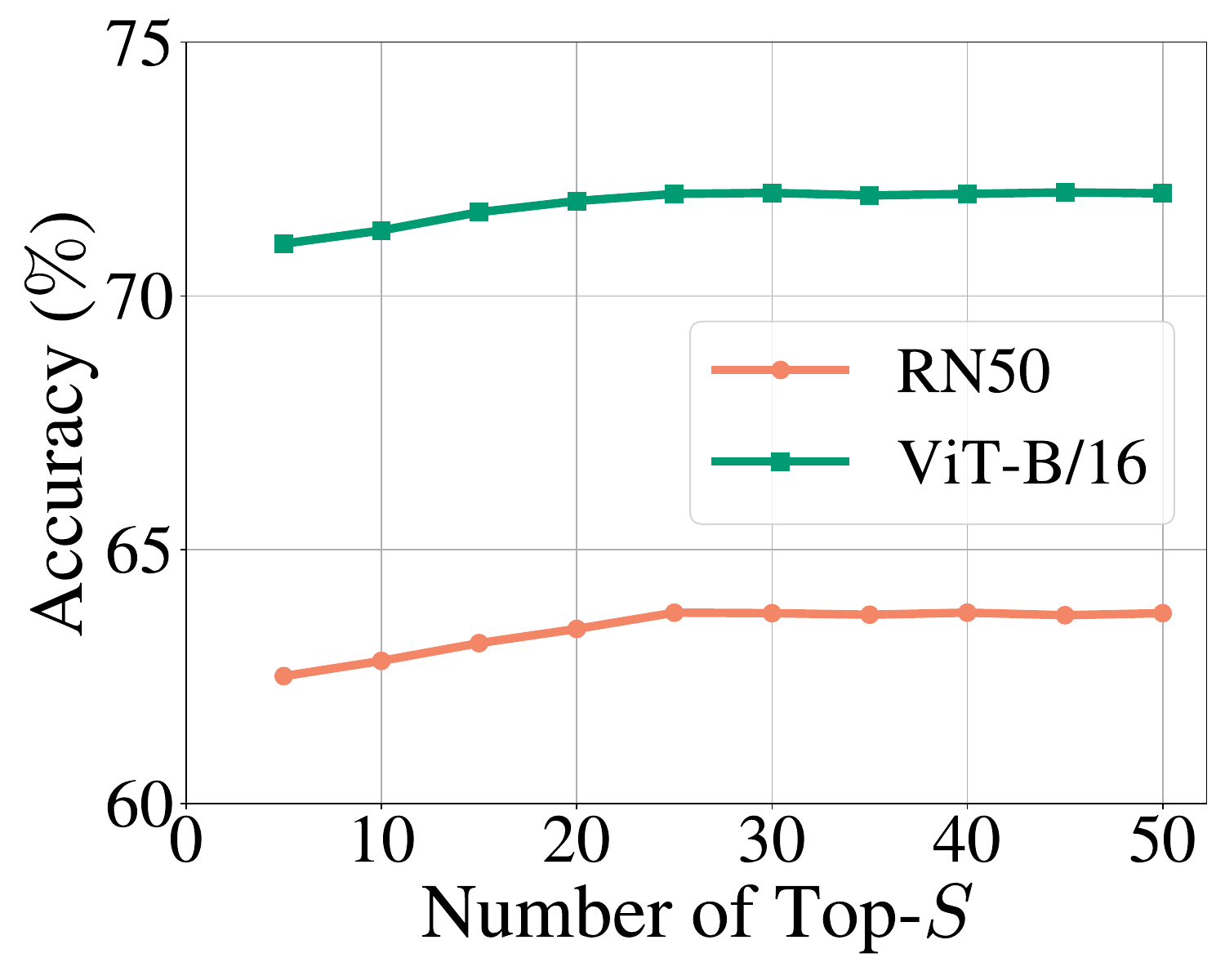}
\caption{Analysis of classification performance by varying the selection number on the ImageNet dataset.} \label{fig:topc}
\end{minipage}
\end{figure}

%% file: figs/heatmap.tex
\begin{figure}
  \centering
    \includegraphics[width=1\linewidth]{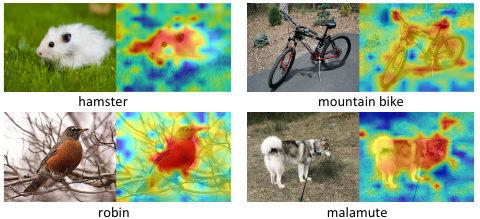}
    \caption{Visualization of heatmaps for the transport plans.}
    \label{fig:visualization}
\end{figure}

%% file: tables/time_consume.tex
\begin{table}[h]
\centering
\tabstyle{5pt}
\begin{tabular}{lccccc}
\toprule
    Method & Testing Time  &Accuracy &Gain\\ \midrule
    CLIP~\cite{CLIP}       &13min &59.81 &0\\
    TPT~\cite{TPT}         &14h\ 30min  &60.74  &+0.93\\
    DiffTPT~\cite{DiffTPT} &37h\ 25min  &60.80  &+0.99\\
    TDA~\cite{TDA}         &17min       &61.35  &+1.54\\
    DOTA~\cite{dota}       &18min  &61.82  &+2.01  \\
    DPE~\cite{DPE}         &2h\ 40min &{63.41} &+3.60 \\
    AWT~\cite{AWT}         &21min &63.01 &+3.20 \\
    \textbf{ProtoMM}          &21min &\textbf{63.76} &\textbf{+3.95} \\
\bottomrule
\end{tabular}
\caption{Comparisons of efficiency (\textit{Testing Time}) and effectiveness (\textit{Accuracy}) with the RN50 backbone. 
}
\label{tab:time}
\end{table}

%% file: section/5_conclusion.tex
\section{Conclusion}
In this work, we introduce a novel training-free multimodal prototype learning framework for adapting VLMs during test time. Our method models prototypes as a discrete distribution of class descriptions and visual particles, where the visual particles are dynamically updated from the testing streams. This process enables the multimodal prototypes to progressively accumulate visual knowledge, enhancing the prediction performance progressively. Extensive experiments demonstrate the effectiveness of our method. 
In the future, we aim to design a more efficient multimodal prototype learning framework, and apply it to more high-level downstream tasks, such as image/video understanding. 

\section{Acknowledge}
This research is supported by the National Natural Science Foundation of China (No.U24B20180) and the advanced computing resources provided by the Supercomputing Center of the USTC. It is also supported by the RIE2025
Industry Alignment Fund – Industry Collaboration Projects
(IAF-ICP) (Award I2301E0026), administered by A*STAR.
